%% file: ijcai24.tex
\documentclass{article}
\usepackage{ijcai24}

\usepackage{times}
\usepackage{soul}
\usepackage{multirow}
\usepackage{url}
\usepackage{soul}
\usepackage[hidelinks]{hyperref}
\usepackage[utf8]{inputenc}
\usepackage[small]{caption}
\usepackage{graphicx}
\usepackage{amsmath}
\usepackage{amsthm}
\usepackage{booktabs}
\usepackage{algorithm}
\usepackage{algorithmic}
\usepackage[switch]{lineno}
\usepackage{amssymb}
\usepackage{color}
\usepackage{makecell}

\newcommand{\myfootnote}[1]{\footnotemark[1]}

\title{TAI++: Text as Image for Multi-Label Image Classification by \\
Co-Learning Transferable Prompt}

\author{
Xiangyu Wu$^{1,2}$
\and
Qing-Yuan Jiang$^{1}$
\and
Yang Yang$^1$\thanks{Corresponding author}\and 
Yi-Feng Wu$^{2}$\and
Qing-Guo Chen$^2$\and \\
Jianfeng Lu$^1$\myfootnote{Corresponding author}
\affiliations
$^1$Nanjing University of Science and Technology\\
$^2$Alibaba International Digital Commerce Group\\
\{wxy\_yyjhl,yyang,lujf\}@njust.edu.cn, qyjiang24@gmail.com, \\
\{yixin.wyf,qingguo.cqg\}@alibaba-inc.com
}

\begin{document}
\maketitle

\begin{abstract}

The recent introduction of prompt tuning based on pre-trained vision-language models has dramatically improved the performance of multi-label image classification. However, some existing strategies that have been explored still have drawbacks, i.e., either exploiting massive labeled visual data at a high cost or using text data only for text prompt tuning and thus failing to learn the diversity of visual knowledge. Hence, the application scenarios of these methods are limited. In this paper, we propose a pseudo-visual prompt~(PVP) module for implicit visual prompt tuning to address this problem. Specifically, we first learn the pseudo-visual prompt for each category, mining diverse visual knowledge by the well-aligned space of pre-trained vision-language models. Then, a co-learning strategy with a dual-adapter module is designed to transfer visual knowledge from pseudo-visual prompt to text prompt, enhancing their visual representation abilities. Experimental results on VOC2007, MS-COCO, and NUSWIDE datasets demonstrate that our method can surpass state-of-the-art~(SOTA) methods across various settings for multi-label image classification tasks. The code is available at \url{https://github.com/njustkmg/PVP}.

\end{abstract}

\section{Introduction}
\label{sec:intro}
\begin{figure}[!t]
	\centering
	\includegraphics[scale=0.7]{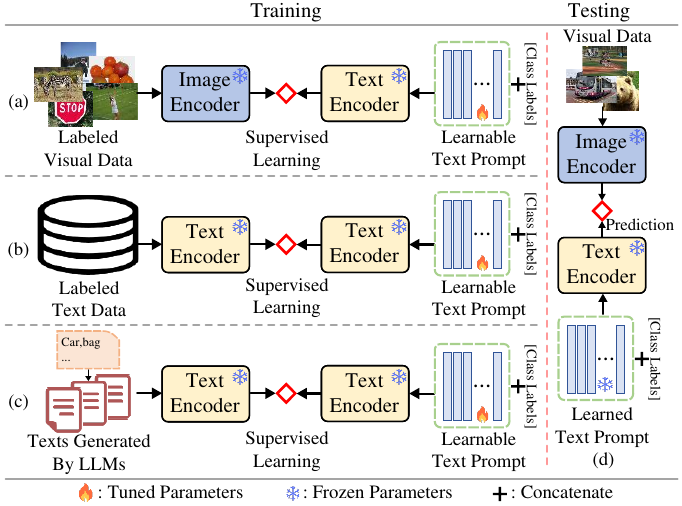}
	\caption{Different prompt tuning paradigms for multi-label image recognition. (a). Tuning text prompt with labeled visual data. (b). Tuning text prompt with labeled text data. (c). Tuning text prompt with texts generated by LLMs. (d). Testing with images and text prompt for image recognition.}
\label{fig:comparsion}
\end{figure}

Multi-label image classification~\cite{MLIC:journals/pami/WeiXLHNDZY16,DualCoOp:conf/nips/SunHS22,CoOp:journals/ijcv/ZhouYLL22,TAIDPT:conf/cvpr/GuoDJBGZ23,LSMM:conf/ijcai/MaoWZ23} has attracted much more attention in many areas including machine learning, computer vision, etc. In recent years, vision-language pre-training models~\cite{CLIP:conf/icml/RadfordKHRGASAM21,ALIGN:conf/icml/JiaYXCPPLSLD21,BridgeTower:conf/aaai/0005WRLCD23,APT:conf/cvpr/BowmanAZTPPS23,BLIP2:conf/icml/0008LSH23,NAIC:conf/aaai/FuSZY24} have exhibited remarkable generalization capabilities by aligning visual and language modalities to shared embedding space. Through utilizing the aligned vision-language embeddings from pre-trained vision-language models, prompt tuning~\cite{CoCoOp:conf/cvpr/ZhouYL022,PPT:conf/acl/GuHLH22,Hierarchical-Prompt:conf/ijcai/WangYWD23,APVT:conf/ijcai/SunLPMZX023} has emerged as a novel paradigm and significantly enhanced performance in multi-label image classification~\cite{CoOp:journals/ijcv/ZhouYLL22,DualCoOp:conf/nips/SunHS22,TAIDPT:conf/cvpr/GuoDJBGZ23}. 

As a pioneering work, CoOp~\cite{CoOp:journals/ijcv/ZhouYLL22}, which is illustrated in Figure~\ref{fig:comparsion}~(a) and Figure~\ref{fig:comparsion}~(d), ingeniously designed learnable text prompt combined with textual class labels, aligned with images in a supervised manner via frozen encoders. CoCoOp~\cite{CoCoOp:conf/cvpr/ZhouYL022} explored a conditional context of image strategy for unseen classes. To further leverage aligned space of origin CLIP~\cite{CLIP:conf/icml/RadfordKHRGASAM21}, DualCoOp~\cite{DualCoOp:conf/nips/SunHS22} encoded dual~(positive and negative) prompts with partial-labeled images. DualCoOp++\cite{DualCoOp++:journals/pami/HuSXSS23} separately encoded evidential, positive, and negative prompts to further improve the performance. All of these methods necessitate labeled visual data for model training. However, constructing adequate labeled visual data is costly. And insufficient training data might hinder the learning of robust image recognition networks. Hence, the application scenarios of these methods are limited.

Considering that the visual and text modalities are well-aligned by pre-trained vision-language models, such as CLIP, TAI-DPT~\cite{TAIDPT:conf/cvpr/GuoDJBGZ23} proposed to enable low-cost text data instead of labeled visual data to learn text prompt. In other words, TAI-DPT leveraged publicly labeled text data for tuning text prompt and directly classified images during testing. The training and testing procedure of TAI-DPT is depicted in Figure~\ref{fig:comparsion}~(b) and Figure~\ref{fig:comparsion}~(d), respectively. Apparently, TAI-DPT failed to learn the diversity of visual knowledge because TAI-DPT only utilized text data and textual class labels for text prompt tuning.

There also exist some more challenging scenarios compared with those mentioned above. A typical scenario is that only labeled text data of common categories is available. In this scenario, a reliable option is to utilize the large language models~(LLMs)~\cite{GPT4:journals/corr/abs-2303-08774,ChatGLM:conf/iclr/ZengLDWL0YXZXTM23} to generate sufficient pseudo text data for prompt tuning. For example, Guangdong-Hong Kong-Macao International Algorithm Competition\footnote{https://iacc.pazhoulab-huangpu.com/} advises using LLMs and textual class labels to generate pseudo texts. We can use generated pseudo text data and adopt the same strategy as TAI-DPT, which only uses text data to perform prompt tuning. The training and testing procedure is illustrated in Figure~\ref{fig:comparsion}~(c) and Figure~\ref{fig:comparsion}~(d), respectively. However, for multi-label image classification, learning a wide range of discriminative visual-level information for each category is essential and indispensable due to the heterogeneity between images and texts. For example, different images with the same categories, like cars, planes, or bags, encompass various shapes and attributes. To sum up, existing prompt tuning based multi-label classification methods either require a large amount of labeled visual data or fail to learn the diversity of visual knowledge if the algorithm solely adopts text data to perform text prompt tuning.

How to reconcile this problem? In this paper, we design a pseudo-visual prompt module for visual prompt tuning implicitly, mining diverse visual knowledge and explicitly avoiding the usage of massive labeled visual data. More concretely, we propose a co-learning method for both visual and text prompt tuning, where the visual prompt tuning is performed implicitly based on the pseudo-visual prompt module. As visual and text modalities are well-aligned by pre-trained models, we can learn diverse visual knowledge from aligned space of CLIP instead of using massive labeled visual data. Then, we co-learn visual and text prompts with a dual-adapter module by transferring visual knowledge from learned pseudo-visual prompt to text prompt. Our contributions are summarized as follows:
\begin{itemize}
    \item We design a pseudo-visual prompt module to perform visual prompt tuning implicitly and propose a novel transferable prompt co-learning method for both visual and text prompt tuning for multi-label image classification. Furthermore, our pseudo-visual prompt can be easily combined with existing methods to improve multi-label image classification performance further.
    \item We co-learn visual and text prompts by leveraging a dual-adapter module and contrastive learning for transferring visual knowledge to text prompt, thereby enhancing their visual representation capabilities.
    \item Extensive experiments on three widely-used benchmarks, i.e., VOC2007, MS-COCO, and NUSWIDE, show that our proposed method can outperform SOTA methods for multi-label image classification tasks.
\end{itemize}

\section{Related Work}

\noindent\textbf{Multi-Label Image Classification}. Given an image input, multi-label image classification~\cite{MLIC:conf/cvpr/WangYMHHX16,LASO:conf/cvpr/AlfassyKASHFGB19,COC:conf/kdd/YangWZL018,CoOp:journals/ijcv/ZhouYLL22,PGLoss:conf/aaai/Lin23,RSSL:conf/icdm/XiSGY23} tries to recognize all the target class labels. Early works~\cite{MLIC:conf/cvpr/WangYMHHX16,MLIC:conf/iccv/WangCLXL17} focus on utilizing labeled visual data to learn intra-class relationships. However, these methods usually cannot achieve satisfactory performance when labeled visual data is limited. Recent works~\cite{MLIC:conf/cvpr/LeeFYW18,LASO:conf/cvpr/AlfassyKASHFGB19,Optimal-Transport:journals/tkde/YangFZLJ21,Meta:conf/wacv/SimonKH22,DPML:conf/ijcai/WangYL0HCFC23} pay more attention to the data-limited scenarios, including the scenarios with zero-shot, few-shot and partial-labeled image data, in past years. 

The partial-labeled data is defined as that some class labels of data are unknown. The authors of~\cite{MLIC:conf/cvpr/LeeFYW18} utilized knowledge graph to mine the relationship between different class labels. SARB~\cite{SARB:conf/aaai/0002CWL22} tried to learn instance-level and prototype-level semantic representations to complement unknown labels. Zero/few-shot multi-label image classification tries to learn novel class labels from limited data. LaSO~\cite{LASO:conf/cvpr/AlfassyKASHFGB19} leveraged relationships among label sets to extract underlying semantic information for few-shot image classification. In paper~\cite{Semantic:journals/corr/abs-2303-14123}, the authors introduced an embedding matrix with principal embedding vectors trained using a tailored loss function.

\begin{figure*}
    \centering
    \includegraphics[scale=1.3]{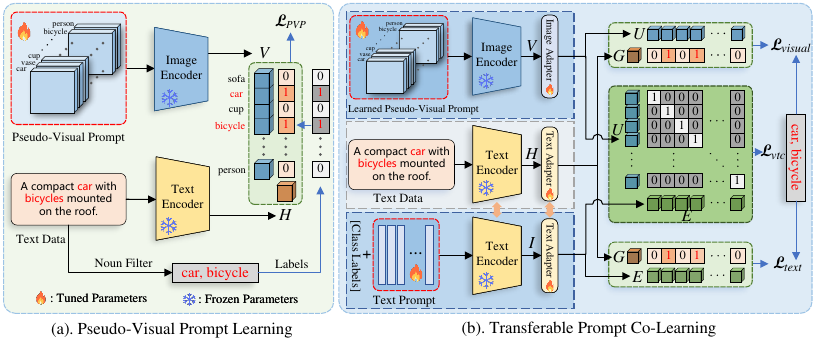}
    \caption{Pseudo-Visual Prompt Learning and Transferable Prompt Co-Learning. Sub-Figure~(a) presents the class-specific pseudo-visual prompt module. The global text embedding and pseudo-visual prompt embedding are obtained from the frozen CLIP image and text encoders. The corresponding cosine similarity between the embeddings is guided by the noun-filtered labels with ranking loss. Sub-Figure~(b) presents the transferable prompt co-learning module. We perform contrastive learning between the pseudo-visual prompt and the text prompt to enhance the prompts' visual diversity representation capability.}
    \label{fig:ourframework}
\vspace{-1em}
\end{figure*}

\noindent\textbf{Prompt Tuning in Multi-Modal Learning}. Prompt tuning~\cite{Noisy:conf/acl/MinSMHZL22,PLPrompt:conf/ijcai/YanHXLW23,DVLF:journals/chinaf/YangBGZYY23,APT:conf/cvpr/BowmanAZTPPS23,Black-box:conf/ijcai/YuCL023} has emerged as a promising technique in computer vision and natural language processing, offering a parameter-efficient way to leverage large pre-trained models. KnowPrompt~\cite{KnowPrompt:conf/www/ChenZXDYTHSC22} involved injecting knowledge into a prompt template and encoding rich semantic knowledge among entities and relations. Pro-Tuning~\cite{ProTuning:journals/corr/abs-2207-14381} learned task-specific visual prompt for downstream input images while keeping the pre-trained model frozen. 

Amidst the progress in multi-modal pre-training, researchers have explored the application of prompt tuning in the multi-modal domain. CoOp~\cite{CoOp:journals/ijcv/ZhouYLL22} modified the pre-trained vision-language models for image recognition tasks by employing learnable prompt context vectors. DualCoOp++~\cite{DualCoOp++:journals/pami/HuSXSS23} efficiently adapted a powerful vision-language model with partial-labeled images by introducing evidence-guided region feature aggregation and winner-take-all modules to improve spatial aggregation and inter-class interaction. These methods necessitated visual modality and textual class labels as default prerequisites in both training and testing. Consequently, TAI-DPT~\cite{TAIDPT:conf/cvpr/GuoDJBGZ23} extended this paradigm by treating texts as images for zero-shot image recognition, storing the aligned vision-language information from CLIP into text prompt without seeing any image during training.

Prompt tuning based methods can boost the performance of multi-label image classification. However, these methods either require a large amount of labeled visual data or fail to learn the diversity of visual knowledge. In this paper, we propose a novel transferable prompt co-learning method to solve this problem by designing a pseudo-visual prompt module.

\section{Methodology}
In this section, we introduce the architecture of our proposed pseudo-visual prompt~(PVP) method in detail. The whole architecture of our proposed method is illustrated in Figure~\ref{fig:ourframework}. Our proposed method comprises two key phases: the pseudo-visual prompt learning phase and the transferable prompt co-learning phase. During training, these two phases will be performed sequentially. In practice, we observed that direct co-learning pseudo-visual prompt and text prompt, without pre-learning pseudo-visual prompt, will lead to difficulty in convergence of pseudo-visual prompt. The reason is that the PVP is initialized randomly without textual class labels while the text prompt combines textual class labels. Hence, we adopt the two-stage learning strategy for our proposed method. Furthermore, we introduce two training text construction strategies for the scenarios of labeled visual data and unavailable text data respectively.
\subsection{Pseudo-Visual Prompt Learning}
\label{sec:PVP}
The accurate learning of diverse and comprehensive visual knowledge for each class label is pivotal for image classification tasks. Text prompt, combined with textual class labels, can only capture visual information aligned with the textual class labels from the aligned space. Hence, we propose a pseudo-visual prompt module, aiming to construct a class-specific visual prompt for each category and leverage the image encoder, text encoder, and aligned space of CLIP to optimize the pseudo-visual prompt, learning the generic visual knowledge for each category.

As shown in Figure~\ref{fig:ourframework}~(a), we innovatively design the class-specific pseudo-visual prompt for each category without combining them with any explicit visual or textual labels. Image modality encompasses a wide range of diversity, which has been learned in the well-aligned space of CLIP. Hence, the class-specific pseudo-visual prompt can learn and store the unique visual knowledge of each category. More concretely, the pseudo-visual prompt is defined as:
\begin{equation}
    {\boldsymbol P} = [\;{\boldsymbol p}_1, {\boldsymbol p}_2, {\boldsymbol p}_3, ..., {\boldsymbol p}_N\;],\label{eq:pvp}
\end{equation}
where ${\boldsymbol p}_{i} \in \mathbb{R} ^{H\times W\times 3 }$ denotes class-specific pseudo-visual prompt for $i$-th class, $H$ and $W$ are equal in size, $N$ is the number of class labels. Note that the number of pseudo-visual prompt is batch-size agnostic and equals the number of target categories. 

Then, we utilize the well-aligned space of CLIP, collected text training data, and frozen image/text encoder to learn pseudo-visual prompt. The learning procedure of pseudo-visual prompt can be formalized as follows: 
\begin{equation}
\langle\Omega_{{aligned}}, \boldsymbol{T}, \phi, \psi\rangle \rightarrow \langle{\boldsymbol P}\rangle,\nonumber
\end{equation}
where $\boldsymbol{T}$ represents the collected labeled text data or pseudo text data generation by LLMs, $\Omega_{aligned}$ represents the origin CLIP's aligned shared space, $\phi(\cdot)$ and $\psi(\cdot)$ refer to the frozen text and image encoders of CLIP, respectively. As for the input text $\boldsymbol{T}$, we directly follow the origin CLIP to obtain the global text embedding by projecting the feature of the last ``${<}\text{EOS}{>}$'' token. The global visual embedding for each category of ${\boldsymbol P}$ is obtained by visual attention pooling. Hence, we have:
\begin{equation}
{\boldsymbol H}=\phi(\boldsymbol{T} ), \quad{\boldsymbol V}=\psi(\boldsymbol{P} ),
\nonumber
\end{equation}
where ${\boldsymbol H} \in \mathbb{R} ^{B\times D} $ denotes the extracted normalized global text embeddings of a batch, and  ${\boldsymbol V} \in \mathbb{N} ^{N\times D} $ denotes the normalized global visual embeddings of $N$ pseudo-visual prompts. For a specific text $\boldsymbol{t}_i\in\boldsymbol{T}$ in a batch, the similarity of text $\boldsymbol{t}_i$ and pseudo-visual prompt can be computed by: 
\begin{equation}\label{eq:e5}
    s_{ij} = \langle {\boldsymbol h}_i,\; {\boldsymbol v}_{j} \rangle,\quad \forall j\in\{1,2,3,...,N\}.
\end{equation}
Here, ${\boldsymbol h}_i\in{\boldsymbol{H}}, {\boldsymbol v}_{j}\in{\boldsymbol{V}}$ denote the global text embedding of text $\boldsymbol{t}_i$ and the global visual embedding of $j$-th pseudo-visual prompt, respectively. We then perform noun filtering to obtain the positive and negative labels. Specifically, given a text embedding $\boldsymbol{h}_i$ and a pseudo-visual prompt embeddings $\boldsymbol{v}_j$, if the class label filtered from input text by noun filtering is contained in the target category set, $\boldsymbol{h}_i$ and $\boldsymbol{v}_j$ are positive pair. Otherwise, they are negative pairs. We employ the ranking loss to measure the discrepancy between similarity scores and text labels following the setting of TAI-DPT~\cite{TAIDPT:conf/cvpr/GuoDJBGZ23}:
\begin{equation}
\mathcal{L}_{{PVP}} = \frac{1}{B} \sum_{k=1}^{B} \sum_{i \in \{c^+ \}} \sum_{j \in \{c^- \}} \max(0, m - s_{ki} + s_{kj}),\label{eq:obj-pvp}
\end{equation}
where $c^+$ and $c^-$ are positive labels and negative labels, $s_{ki}$ and $s_{kj}$ are positive pair and negative pair similarities described in Equation~(\ref{eq:e5}), $m$ is the margin used to measure the difference between each pair of predicted values. During training, we fix the text encoder and image encoder and only learn the pseudo-visual prompt by optimizing the objective function in Equation~(\ref{eq:obj-pvp}).

\subsection{Transferable Prompt Co-Learning}
After the first learning phase, the diverse class-specific visual knowledge is well aligned with class labels and stored in pseudo-visual prompt. Furthermore, inspired by TAI-DPT~\cite{TAIDPT:conf/cvpr/GuoDJBGZ23}, we design contrastive loss and a dual-adapter module to co-learn the pseudo-visual prompt and text prompt by transferring visual information to the text prompt. Here, the pseudo-visual prompt is initialized by the learned pseudo-visual prompt in the first phase.

Figure~\ref{fig:ourframework}~(b) illustrates the transferable prompt co-learning procedure. We first adopt the same definition of pseudo-visual prompt $\boldsymbol{P}$ from Equation~(\ref{eq:pvp}) during this phase. Then, we define the text prompt as follows:
\begin{equation}
\begin{split}
&\forall i\in\{1,2,\dots,N\},\;\boldsymbol{R}_i = [\;\boldsymbol{r}_1, \boldsymbol{r}_2, ...,\boldsymbol{r}_M, \boldsymbol{g}_i\;],\nonumber\\
&\boldsymbol{S}=[\;\boldsymbol{R}_1,\boldsymbol{R}_2,\dots,\boldsymbol{R}_N\;],
\end{split}
\end{equation}
where ${\boldsymbol{g}}_i$ denotes word embedding of the $i$-th class label, $\boldsymbol{r}_i$ is a learnable context vector of text prompt and $M$ denotes the number of text prompt.

Then, we utilize image and text encoders from CLIP to obtain pseudo-visual prompt, text prompt and global text embeddings.  
\begin{equation}
{\boldsymbol{V}}=\psi(\boldsymbol{P}),\;{\boldsymbol{H}}=\phi(\boldsymbol{T}),\;{\boldsymbol{I}}=\phi(\boldsymbol{S}).\nonumber
\end{equation}
To mine the knowledge of origin CLIP and downstream task, we apply an identical adapter module~\cite{CLIPAdapter:journals/corr/abs-2110-04544} for image encoder and text encoder, a.k.a., a dual-adapter module. Both image and text adapter consist of two fully connected layers, an activation function, and a residual connection. We apply an image adapter for pseudo-visual prompt and a text adapter for both text prompt and global text embedding extraction. Then, we have:
\begin{equation}\label{eq:dual-adapter}
\begin{split}
&{\boldsymbol{U}}=(1-\lambda )g({\boldsymbol{V}}) +\lambda {\boldsymbol{V}},\\
&{\boldsymbol{G}}=(1-\lambda )h({\boldsymbol{H}}) +\lambda {\boldsymbol{H}},\\
&{\boldsymbol{E}}=(1-\lambda )h({\boldsymbol{I}}) +\lambda {\boldsymbol{I}}.
\end{split}
\end{equation}
Here, $g(\cdot)$ and $h(\cdot)$ denote the adapter functions for image and text modality, respectively. And $\lambda\in[0,1]$ denotes the weight between feature from adapter module and feature from image/text encoder. Please note that text prompts and global text utilize the same adapter in Equation~(\ref{eq:dual-adapter}), which means text adapter is a parameter-sharing network for text prompt and global text learning.

After embedding extraction, we design the objective function for pseudo-visual prompt and text prompt co-learning. We present the objective function based on the given pseudo-visual prompt, text prompt and global text embeddings. Specifically, we first utilize contrastive loss to preserve the similarity between pseudo-visual prompt and text prompt of $N$ class labels. The similarity matrix can be obtained by: $\boldsymbol{U}\boldsymbol{E}^T\in\mathbb{R}^{N\times N}$. And the ground truth for pseudo-visual prompt and text prompt of $N$ class labels is an identity matrix. Note that the size of similarity matrix is batch-size agnostic and equals the number of target categories $N$. Thus, the contrastive loss can be written as:
\begin{equation}\nonumber
\begin{split}
&p_{ij}^{v2t}=\frac{\exp{\big(s(\boldsymbol{u}_i,\boldsymbol{e}_j)/\tau\big)}}{\sum_{k=1}^N\exp{\big(s(\boldsymbol{u}_i,\boldsymbol{e}_k)/\tau\big)}},\\
&p_{ij}^{t2v}=\frac{\exp{\big(s(\boldsymbol{u}_i,\boldsymbol{e}_j)/\tau\big)}}{\sum_{k=1}^N\exp{\big(s(\boldsymbol{u}_k,\boldsymbol{e}_j)/\tau\big)}},\\
&\mathcal{L}_{{vtc}}=\frac{1}{2}\big[l_{CE}(y^{{v2t}},p^{{v2t}})+l_{{CE}}(y^{{t2v}},p^{{t2v}})\big],
\end{split}
\end{equation}
where $p_{ij}^{v2t}$ and $p_{ij}^{t2v}$ denote the softmax-normalized similarity from pseudo-visual prompt to text prompt and from text prompt to pseudo-visual prompt, respectively. $\tau$ denotes the temperature scale parameter. $l_{{CE}}$ denotes cross-entropy loss. $\forall i,j\in\{1,2,\dots,N\},y_{ij}^{v2t},y_{ij}^{t2v}\in\{0,1\}$ denote the similarity ground-truth of text prompt and pseudo-visual prompt. $y_{ij}^{v2t}=1$ if text prompt $\boldsymbol{t}_i$ and pseudo-visual prompt $\boldsymbol{p}_j$ belong to the same category, and $y_{ij}^{v2t}=0$ otherwise. The definition of $y_{ij}^{t2v}$ is the same with $y_{ij}^{v2t}$.

Moreover, we utilize ranking loss to obtain $\mathcal{L}_{{visual}}$ and $\mathcal{L}_{{text}}$ similar to Equation~(\ref{eq:obj-pvp}). Specifically, we use $\mathcal{L}_{{visual}}$ to measure the disparity between global text and pseudo-visual prompt. The $\mathcal{L}_{{visual}}$ can be formulated as follows:
\begin{align}
&s^{v}_{ij}=\langle {\boldsymbol g}_i,\; {\boldsymbol u}_{j} \rangle,\;\forall j\in\{1,2,3,...,N\},\nonumber\\
&\mathcal{L}_{{visual}}=\frac{1}{B} \sum_{k=1}^{B} \sum_{i \in \{c^+ \}} \sum_{j \in \{c^- \}} \max(0, m - s^{v}_{ki} + s^{v}_{kj}).\nonumber
\end{align}

Similarly, the $\mathcal{L}_{text}$ that measures the disparity between global text and text prompt can be formulated as follows:
\begin{align}
&s^{t}_{ij}=\langle {\boldsymbol g}_i,\; {\boldsymbol e}_{j} \rangle,\;\forall j\in\{1,2,3,...,N\},\nonumber\\
&\mathcal{L}_{{text}}=\frac{1}{B} \sum_{k=1}^{B} \sum_{i \in \{c^+ \}} \sum_{j \in \{c^- \}} \max(0, m - s^{t}_{ki} + s^{t}_{kj}).\nonumber
\end{align}

Finally, we get the total training loss by combining $\mathcal{L}_{vtc}$, $\mathcal{L}_{{visual}}$ and $\mathcal{L}_{{text}}$:
\begin{equation}
\mathcal{L}=\mathcal{L}_{{vtc}}+\mathcal{L}_{{visual}}+\mathcal{L}_{{text}}.\label{eq:co-learn-obj}
\end{equation}

During training procedure, we fix the image and text encoder from CLIP and learn pseudo-visual prompt, text prompt and dual-adapter by optimizing the objective function in Equation~(\ref{eq:co-learn-obj}).

\subsection{Training Text Data Construction}
In this section, we discuss the training text data construction for different application scenarios. To obtain training text data in this paper, we utilize two strategies, i.e., human-annotated labeled text data construction~\cite{TAIDPT:conf/cvpr/GuoDJBGZ23} and LLMs-based pseudo text data construction. The first strategy was introduced by TAI-DPT~\cite{TAIDPT:conf/cvpr/GuoDJBGZ23}, and we follow the setting of TAI-DPT. Specifically, we directly employ public object detection datasets like MS-COCO~\cite{COCO:conf/eccv/LinMBHPRDZ14} and localized narratives~\cite{OpenImages} to form labeled text data. For the second strategy, pseudo text data is generated using constructed templates and LLMs~\cite{GPT4:journals/corr/abs-2303-08774,ChatGLM:conf/iclr/ZengLDWL0YXZXTM23} for automatic generation. Concretely, we first combine several class labels with a query template to construct a query prompt. Then, we utilize LLMs to generate pseudo text data. We provide a query prompt example as follows:

\noindent\textit{{PROMPT}: Make a sentence to describe a photo. Requirements: Each sentence should be less than 15 words and include keywords: car, dog, cat.}

To filter out the unreliable texts generated by LLMs, we re-input the generated text data into LLMs with another query template: 

\noindent\textit{PROMPT: Will the scene described in this text appear in reality? Scene: + ``\{text\}''}.

Moreover, we judge the reasonableness of the text through the output \textit{likely/unlikely}. For word-level filtered labels in input text, we follow the setting of TAI-DPT~\cite{TAIDPT:conf/cvpr/GuoDJBGZ23} using NLTK\footnote{https://www.nltk.org/} to perform noun filtering. More details of the query prompts, examples and noun filtering are provided in the supplementary materials.

\begin{table*}[ht]
\centering
\begin{minipage}{.6\textwidth}
\centering
\caption{The mAP results for zero-shot setting on all datasets. The best performance is shown in boldface. (480) denotes the image resolution during inference}
\begin{tabular}{l|c|c|c|c|c|c}
\Xcline{1-7}{0.75pt}
\multirow{2}{*}{{Method}} & \multicolumn{2}{c|}{{VOC2007}} & \multicolumn{2}{c|}{{MS-COCO}} & \multicolumn{2}{c}{{NUSWIDE}} \\
\cline{2-7}
 & {Label} & {Pseudo} & {Label} & {Pseudo} & {Label} & {Pseudo} \\
\hline
ZSCLIP & \multicolumn{2}{c|}{77.3} & \multicolumn{2}{c|}{49.7} & \multicolumn{2}{c}{37.4} \\
TAI-DPT & \multicolumn{1}{c}{88.3} &  88.1 & \multicolumn{1}{c}{65.1} & 64.6 & \multicolumn{1}{c}{46.5} & 47.3 \\
{PVP} & \multicolumn{1}{c}{\textbf{88.6}} & \textbf{88.9} & \multicolumn{1}{c}{\textbf{67.7}} & \textbf{67.5} & \multicolumn{1}{c}{\textbf{47.6}} & \textbf{49.3} \\
\hline
TAI-DPT~(480) & \multicolumn{1}{c}{88.3} & 88.4 & \multicolumn{1}{c}{67.2} & 66.6 & \multicolumn{1}{c}{42.9} & 44.1 \\
{PVP~(480)} & \multicolumn{1}{c}{\textbf{89.7}} & \textbf{90.0} & \multicolumn{1}{c}{\textbf{70.9}} & \textbf{70.8} & \multicolumn{1}{c}{\textbf{44.3}} & \textbf{46.0} \\
\Xcline{1-7}{0.75pt}
\end{tabular}
\label{tab:zero-shot}
\end{minipage}\hspace{0.02\textwidth}%
\begin{minipage}{.35\textwidth}
\centering
\caption{The mAP results for comparison with few-shot methods on MS-COCO dataset.}
\vspace{-1em}
\begin{tabular}{l|ccc}
\Xcline{1-4}{0.75pt}
{Method} & {0-shot} & {1-shot} & {5-shot} \\
\hline
LaSO & - & 45.3 & 58.1 \\
ML-FSL & - & 54.4 & 63.6 \\
KGGR & - & 52.3 & 63.5 \\
NLC & - & 56.8 & 64.8 \\
TAI-DPT & 59.2 & - & - \\
{PVP} & 62.1 & - & - \\
{PVP~(480)} & \textbf{64.4} & - & - \\
\Xcline{1-4}{0.75pt}
\end{tabular}
\label{tab:few-shot}
\end{minipage}
\end{table*}

\begin{figure*}[ht]
\centering
\includegraphics[scale=0.9]{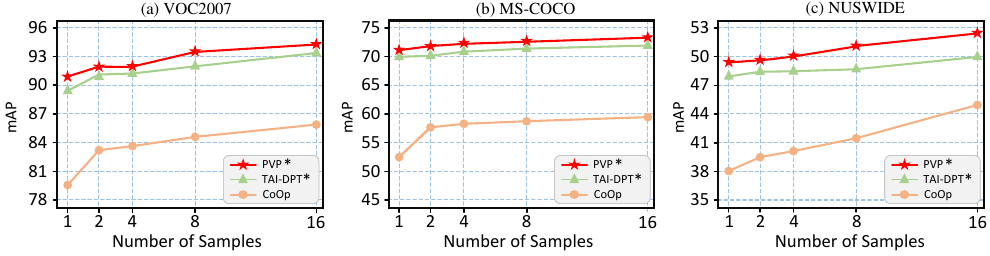}
\caption{Results for few-shot setting, where the performance of PVP*/TAI-DPT* integrate the predictions of PVP/TAI-DPT and CoOp.}
\label{fig:few-shot}
\end{figure*}

\subsection{Model Inference}
During testing procedure, we first replace the input text data with testing images and utilize the image encoder of CLIP to obtain the image embeddings. Then, we utilize image embeddings to compute visual and textual cosine similarities with class label embedding generated by the pseudo-visual prompt and the text prompts, respectively. The final classification score is obtained by fusing the visual and textual cosine similarity.

\section{Experiments}
\subsection{Experiment Setup}

\noindent \textbf{Datasets.} We evaluate our proposed PVP on VOC2007~\cite{VOC2007:journals/ijcv/EveringhamGWWZ10}, MS-COCO~\cite{COCO:conf/eccv/LinMBHPRDZ14}, and NUSWIDE~\cite{NUSWIDE:conf/civr/ChuaTHLLZ09} datasets. The VOC2007 dataset contains 20 categories with 5,011 images for training and 4,952 images for testing. The MS-COCO dataset contains 80 categories divided into training, testing, and validation sets. We use its training set~(82,081 images) for training and validation set~(40,504 images) for testing in our experiments because the class labels of origin testing are unavailable. The NUSWIDE dataset contains 81 categories with 161,789 training images and 107,859 testing images to validate our method. 

To construct the labeled training text data, we follow the setting of TAI-DPT~\cite{TAIDPT:conf/cvpr/GuoDJBGZ23}. More concretely, we extract 100K coco-captions as the text training data for VOC2007 and MS-COCO, and localized narratives from OpenImages~\cite{OpenImages} for NUSWIDE. For the pseudo text data generated by LLMs, we adopt ChatGLM~\cite{ChatGLM:conf/iclr/ZengLDWL0YXZXTM23} to generate 500k pseudo texts for pseudo-visual prompt learning, and 200k for transferable prompt co-learning. Then we use an identical query prompt and the different number of categories for different datasets, i.e., 20 categories for VOC2007 dataset, 80 categories for MS-COCO dataset, and 81 categories for NUSWIDE dataset.

\noindent \textbf{Implementation Details.} For fair comparison, we follow the setting of TAI-DPT~\cite{TAIDPT:conf/cvpr/GuoDJBGZ23} to choose CLIP ResNet-50~\cite{ResNet:conf/cvpr/HeZRS16} as image encoder and the corresponding CLIP Transformer~\cite{Transformer:conf/nips/VaswaniSPUJGKP17} as text encoder. We adopt SGD algorithm to perform prompt learning for both two phases. The margin is set to be 1 for the ranking loss in both two phases.

For the pseudo-visual prompt learning, we initialize an identical class-specific visual prompt of size $224 \times 224 \times 3$ for each category. All to be learned prompts are randomly initialized by the same mean and standard value, i.e., $mean=0, std=0.02$. To perform pseudo-visual prompt learning, both visual and text encoders are frozen, and only prompts are optimized. The training epoch is set to be 40 for all datasets. The learning rate is empirically initialized with 0.1 and decayed through cosine annealing during training.

For the transferable prompt co-learning, we use the learned PVP in the first phase to initialize the pseudo-visual prompt. We follow the setting of TAI-DPT~\cite{TAIDPT:conf/cvpr/GuoDJBGZ23} to initialize text prompt by randomly sampling from a Gaussian distribution with mean being 0 and variance being 0.02, and the length of text prompts is set to 16. In this phase, the image and text encoders are also frozen while the pseudo-visual prompt, text prompt, and dual-adapter module are optimized. The hyper-parameter $\tau$ is set to be $0.02$. The training epoch is set to 20 for all datasets. The learning rate is set to be 1e-4 and 1e-6 for text prompt and pseudo-visual prompt learning and decay by cosine annealing, respectively. The hyper-parameter $\lambda$ of dual-adapter module is set to be 0.5.

\subsection{Comparison with SOTA Methods}

\noindent \textbf{Results on Zero-Shot Task.} To validate the effectiveness of our proposed pseudo-visual prompt, we compare its performance with zero-shot CLIP~(ZSCLIP)~\cite{CLIP:conf/icml/RadfordKHRGASAM21} and the current SOTA method\footnote{Comparison results with recent work TAI-Adapter~\cite{TAI-Adapter:journals/corr/abs-2312-04160} submitted to arXiv are reported in supplementary materials.} TAI-DPT~\cite{TAIDPT:conf/cvpr/GuoDJBGZ23} on all datasets. Table~\ref{tab:zero-shot} presents the results of our proposed method and baselines on all datasets. In Table~\ref{tab:zero-shot}, ``Label'' and ``Pseudo'' denote the training with labeled text data (e.g. coco-caption~\cite{COCO:conf/eccv/LinMBHPRDZ14}, localized narratives~\cite{OpenImages}) and pseudo text data generated by LLMs, respectively. ``(480)'' denotes that the image resolution is set to be $480\times 480$ during inference.

From Table~\ref{tab:zero-shot}, we can see that our proposed method can achieve the best zero-shot multi-label image recognition performance on all three datasets in all cases. On MS-COCO dataset trained with label text and pseudo text data, our method significantly outperforms TAI-DPT by a large margin of 3.7\% and 4.2\% points respectively on $480 \times 480$ resolution. On VOC2007 and NUSWIDE datasets, our method also improves by 1$\sim$2\% points over TAI-DPT on both labeled and pseudo text data. The results demonstrate that our method is without relying on any labeled visual and text data, making it more valuable for real-world application scenarios. Notably, we found that on NUSWIDE dataset, all methods perform better on pseudo data than labeled data. This might be due to the texts generated by the LLMs being closer to the original training data of CLIP than the localized narrative~\cite{OpenImages}. Moreover, the performance on NUSWIDE dataset with higher image resolution is worse than that with lower image resolution. This is due to the resolution of testing images are smaller~(about $200\times 200$) than 480.

\noindent \textbf{Results on Few-Shot Task.} For few-shot task, we select LaSO~\cite{LASO:conf/cvpr/AlfassyKASHFGB19}, ML-FSL~\cite{MLIC:conf/cvpr/MisraZMG16} and TAI-DPT as baselines and compare the performance of these methods. The LaSO and ML-FSL are used for few-shot setting. Meanwhile, TAI-DPT and our proposed method are used for zero-shot setting. LaSO~\cite{LASO:conf/cvpr/AlfassyKASHFGB19} and ML-FSL~\cite{MLIC:conf/cvpr/MisraZMG16} require labeled visual data for training. Hence, for the few-shot setting, we select 64 categories in MS-COCO as normal classes, and the remaining 16 (bicycle, boat, stop sign, bird, backpack, frisbee, snowboard, surfboard, cup, fork, spoon, broccoli, chair, keyboard, microwave, and vase) as novel classes. For few-shot task, the results are reported in Table~\ref{tab:few-shot}. From Table~\ref{tab:few-shot}, we can find that our method outperforms TAI-DPT by a large margin of 5.2\% points. Furthermore, our proposed method also surpasses ML-FSL trained on 5-shot samples by 0.8\% points.

Furthermore, following the same setting of TAI-DPT~\cite{TAIDPT:conf/cvpr/GuoDJBGZ23}, we randomly sample 1, 2, 4, 8, and 16 samples for each class to train the model, and integrate the predictions conveniently with CoOp~\cite{CoOp:journals/ijcv/ZhouYLL22}. The results are reported in Figure~\ref{fig:few-shot}. In Figure~\ref{fig:few-shot}, ``PVP*'' denotes that the performance is calculated by integrating the predictions of CoOp and our PVP when testing. The definition of ``TAI-DPT*'' is similar to ``PVP*''. From Figure~\ref{fig:few-shot}, we can find that our proposed method achieves the best performance on various few-shot settings on all datasets without seeing any labeled visual data.

\begin{table*}[ht]
\centering
\caption{The mAP results for partial-label setting on all datasets, where the performance of PVP*/TAI-DPT* integrates the predictions of PVP/TAI-DPT and DualCoOp. The best performance is shown in boldface.}
\begin{tabular}{l|l|lllllllll|l}
\Xcline{1-12}{0.75pt}
{Datasets} & {Method} & {10\%} & {20\%} & {30\%} & {40\%} & {50\%} & {60\%} & {70\%} & {80\%} & {90\%} & {Avg.} \\ \hline
\multirow{4}{*}{VOC2007} & SARB & 83.5 & 88.6 & 90.7 & 91.4 & 91.9 & 92.2 & 92.6 & 92.8 & 92.9 & 90.7 \\
 & DualCoOp & 91.4 & 93.8 & 93.8 & 94.3 & 94.6 & 94.7 & 94.8 & 94.9 & 94.9 & 94.1 \\
 & TAI-DPT* & 93.3 & \textbf{94.6} & \textbf{94.8} & 94.9 & \textbf{95.1} & 95.0 & 95.1 & \textbf{95.3} & \textbf{95.5} & 94.8 \\
 & {PVP*} & \textbf{93.7} & 94.4 & 94.7 & \textbf{95.1} & \textbf{95.1} & \textbf{95.2} & \textbf{95.2} & \textbf{95.3} & 95.3 & \textbf{94.9} \\
\hline
\multirow{4}{*}{MS-COCO} & SARB & 71.2 & 75.0 & 77.1 & 78.3 & 78.9 & 79.6 & 79.8 & 80.5 & 80.5 & 77.9 \\
 & DualCoOp & 81.0 & 82.3 & 82.9 & 83.4 & 83.5 & 83.9 & 84.0 & 84.1 & 84.3 & 83.3 \\
 & TAI-DPT* & 81.5 & 82.6 & \textbf{83.3} & \textbf{83.7} & \textbf{83.9} & 84.0 & 84.2 & 84.4 & 84.5 & 83.6 \\
 & {PVP*} & \textbf{81.8} & \textbf{82.8} & \textbf{83.3} & 83.6 & \textbf{83.9} & \textbf{84.1} & \textbf{84.3} & \textbf{84.6} & \textbf{84.8} &\textbf{83.7} \\
\hline
\multirow{3}{*}{NUSWIDE} & DualCoOp & 54.0 & 56.2 & 56.9 & 57.4 & 57.9 & 57.9 & 57.6 & 58.2 & 58.8 & 57.2 \\
 & TAI-DPT* & 56.4 & 57.9 & 57.8 & 58.1 & 58.5 & 58.8 & 58.6 & 59.1 & 59.4 & 58.3 \\
 & {PVP*} & \textbf{56.9} & \textbf{58.4} & \textbf{58.9} & \textbf{59.3} & \textbf{59.5} & \textbf{59.7} & \textbf{59.9} & \textbf{60.1} & \textbf{60.2} & \textbf{59.2} \\ 
\Xcline{1-12}{0.75pt}
\end{tabular}
\label{partial-labeled}
\end{table*}

\noindent \textbf{Results on Partial-Label Task.} For partial-label task, we select SARB~\cite{SARB:conf/aaai/0002CWL22} and DualCoOp~\cite{DualCoOp:conf/nips/SunHS22} as baselines. Following the setting of TAI-DPT~\cite{TAIDPT:conf/cvpr/GuoDJBGZ23}, we use visual training sets with different proportions to complete the training of the SARB and DualCoOp methods. For our proposed method and TAI-DPT, we integrate the predictions of these methods with DualCoOp. The results are shown in Table~\ref{partial-labeled}. We can see that PVP* can achieve higher performances than TAI-DPT* in most cases on all datasets after further integrating PVP and TAI-DPT with DualCoOp.

\begin{figure}[ht]
  \centering
  \includegraphics[trim=0 0 0 0, clip, width=0.475\textwidth]{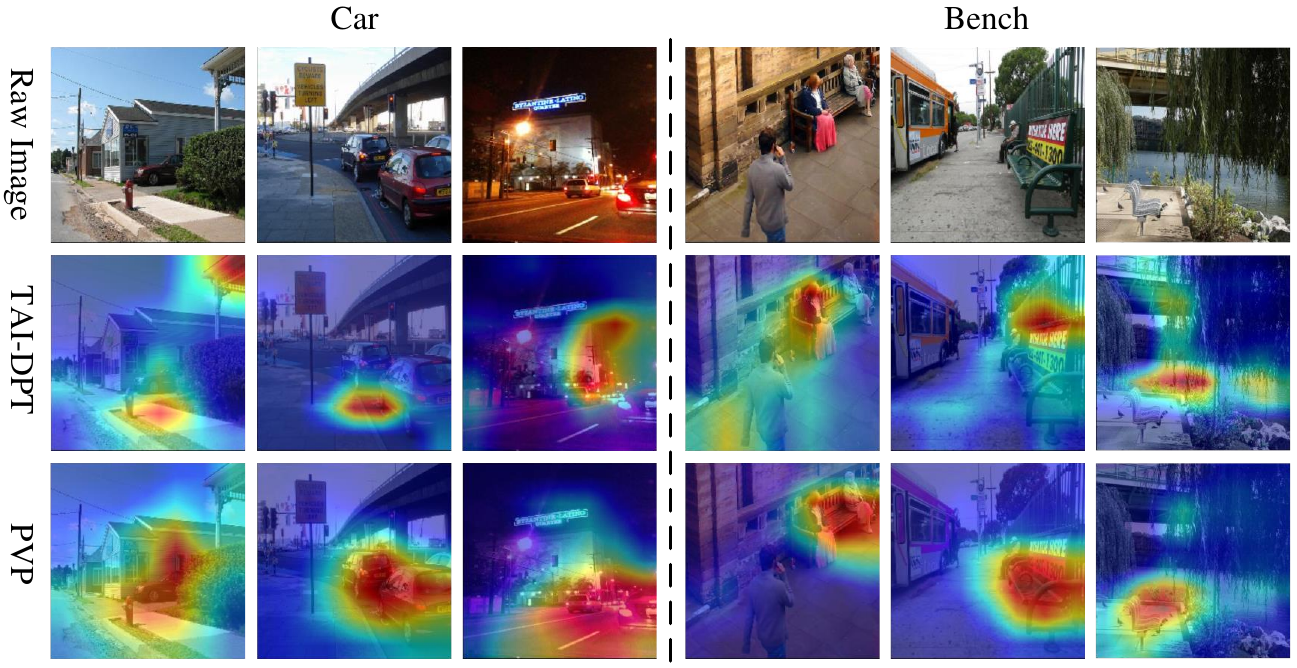}
  \caption{Visualization of PVP and TAI-DPT methods.}\vspace{-10pt}
  \label{fig:visualization}
\end{figure}

\subsection{Visualization} To validate the effectiveness of our proposed method, we conduct several visualization experiments, demonstrating that pseudo-visual prompt are better at focusing on diverse visual information. Specifically, we first select some common class labels, such as car and bench. Then, we randomly select several images of these class labels with different attributes, such as shape, color, size, etc. We then visualize the correlation between the class prompt embeddings of the PVP and TAI-DPT methods and the local image features. The results are shown in Figure~\ref{fig:visualization}. From Figure~\ref{fig:visualization}, we can find that compared with TAI-DPT, PVP can learn extensive features with higher relevance to the class label in different scenes where the object has different shapes or different attributes, and can accurately identify the position of the object even in dark light, bright light or occlusion scenarios. Hence, we demonstrate that our method can learn diverse and comprehensive visual knowledge for each category through pseudo-visual prompt. More visualization results are provided in the supplementary materials.

\subsection{Further Analysis}
\noindent\textbf{Ablation Study.} To evaluate the effectiveness of our methods, we study the influence of different components, including pseudo-visual prompt, dual-adapter, and contrastive learning~(respectively abbreviated as PVP, DA, cLoss in Table~\ref{tab:ablation-study-more}). From Table~\ref{tab:ablation-study-more}, we can observe that all components can improve the performance and the PVP module can boost the most significant improvement.

\begin{figure}[!t]
\centering
\includegraphics[scale=0.75]{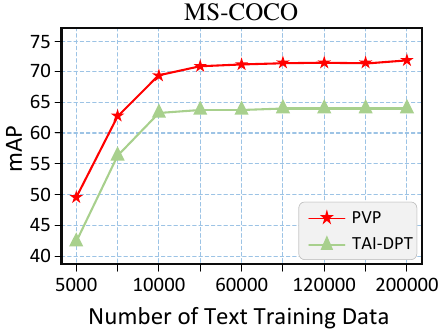}
\caption{The mAP value with different number of text data on MS-COCO dataset.}\vspace{-10pt}
\label{fig:text-data}
\end{figure}

\noindent\textbf{Sensitivity to Hyper-Parameter.} In addition, we further explore the impact of the quantity of text training data on the performance of our method on MS-COCO dataset. To eliminate the influence of different data sources, we mix labeled text and pseudo text data generated by the LLMs and randomly sample different quantities of text as the training set, as shown in Figure~\ref{fig:text-data}. From Figure~\ref{fig:text-data}, we can see that as the number of sampled text data increases, the mAP result increases at the beginning and then remains unchanged. In our experiment, we set the number of text data as 200K. More experimental results can be found in supplementary materials.

\begin{table}[!t]
\centering
\setlength{\tabcolsep}{3.5pt} 
\caption{Ablation study for our proposed method.}
\label{tab:ablation-study-more}
\begin{tabular}{c|c|c|c|c|c}
\Xcline{1-6}{0.75pt}
{PVP} & {DA} & {cLoss} & {VOC2007} & {MS-COCO} & {NUSWIDE} \\ \hline
$\times$ & $\times$ & $\times$ & 88.1 & 64.2 & 47.0 \\
$\checkmark$ & $\times$ & $\times$ & 89.1 & 68.6 & 48.8 \\
$\checkmark$ & $\checkmark$ & $\times$ & 89.4 & 69.7 & 48.9 \\
$\checkmark$ & $\times$ & $\checkmark$ & 89.7 & 69.9 & 48.9 \\
$\checkmark$ & $\checkmark$ & $\checkmark$ & \textbf{90.0} & \textbf{70.8} & \textbf{49.3} \\ 
\Xcline{1-6}{0.75pt}
\end{tabular}
\end{table}

\section{Conclusion}
In this paper, we design a novel pseudo-visual prompt module based on pre-trained vision-language models for multi-label image classification tasks. Thus, we can learn diverse visual knowledge from aligned space of CLIP instead of using massive labeled visual data. By leveraging a contrastive loss and dual-adapter module to co-learn the visual and text prompts, our proposed method can enhance the visual representation capabilities. Experiments verify that our PVP method can achieve the best performance compared with the SOTA methods across various datasets.

\bibliographystyle{named}
\bibliography{ijcai24}

\newpage
\input{Supplementary}

\end{document}

%% file: supplementary.tex











\appendix
\maketitle

\renewcommand{\thesection}{\Alph{section}}

\section{Appendix Overview}
Here we provide more information about our proposed PVP, more ablation studies, and visualization results. The appendix is organized as follows. In Appendix B, we introduce the construction of text training data in detail and present examples of generated text data and synonym dictionaries. In Appendix C, we compare our method with the recent work TAI-Adapter with different LLMs of text generation. We further present more visualization results on the MSCOCO dataset in Appendix D and conduct hyper-parameter experiments in Appendix E.

\section{Training Text Data Construction}
We present the details of the training text data construction in this section.

\begin{figure}[!ht]
	\centering
	\includegraphics[width=\linewidth]{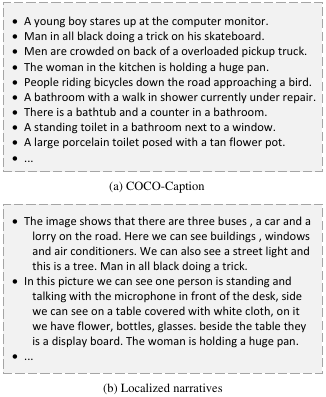}
	\caption{Labeled text data constructed from COCO-Caption of MS-COCO and localized narratives of OpenImages.}
\label{fig:labeled-text}
\end{figure}

\noindent{\textbf{Human-Annotated Labeled Text Data Construction}}.
For VOC2007~\cite{VOC2007:journals/ijcv/EveringhamGWWZ10} and MS-COCO~\cite{COCO:conf/eccv/LinMBHPRDZ14} benchmarks, we obtain the labeled coco-captions from MS-COCO, each text succinctly describes a natural scene, with a maximum length of 25. For NUSWIDE~\cite{NUSWIDE:conf/civr/ChuaTHLLZ09} dataset, we collect localized narratives from the OpenImages~\cite{OpenImages}. Each text contains detailed content descriptions, with a maximum length of 60. The examples are shown in Figure \ref{fig:labeled-text}.


\begin{figure}[!htb]
	\centering
	\includegraphics[width=\linewidth]{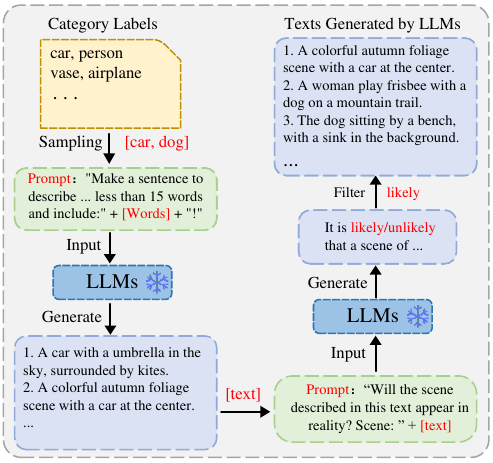}
	\caption{Pseudo text generated by LLMs. Via labels sampling and rationality judgment, obtaining semantically reasonable sentences.}
\label{fig:text-generate}
\end{figure}

\begin{table*}[t]
\caption{Comparison with TAI-Adapter on all datasets. The best performance is shown in boldface.}
\label{tab:tai-adapter}
\centering
\renewcommand{\arraystretch}{1.3} 
\setlength{\tabcolsep}{10pt} 
\begin{tabular}{c|c|c|c|c}
\toprule[0.8pt]
{Method}     & LLM         &  VOC2007 & MS-COCO & NUSWIDE \\ \hline
TAI-DPT      & N/A         &  88.3    & 65.1    & 46.5    \\
TAI-Adapter  & Vicuna-33b  &  89.0    & 67.7    & \textbf{53.3}    \\
PVP          & ChatGLM     &  88.9    & 67.5    & 49.3    \\
PVP          & Vicuna-33b  &  \textbf{89.5}    & \textbf{68.9}    & 51.4    \\
\bottomrule[0.8pt]
\end{tabular}
\end{table*}

\noindent{\textbf{LLMs-based Pseudo Text Data Construction.}} We describe the LLMs-based pseudo text data construction in detail. Figure~\ref{fig:text-generate}  illustrates the process of LLMs generating pseudo text data through artificially designed prompt templates and a set of category labels. Given a target category set $\mathcal{C}=\{c_1, c_2, \dots, c_N\}$, where $N$ denotes the number of categories and $c_i$ denotes a particular class, we design a query prompt as follows: 

\noindent\textit{PROMPT: Make a sentence to describe a photo. Requirements: Each sentence should be less than 15 words and include keywords: $\{c_{i_1}, c_{i_2}, \dots, c_{i_l}\}$}.

\noindent Here $\{c_{i_1}, c_{i_2}, \dots, c_{i_l}\} \subset \mathcal{C}$ and $l\leq 3$. Then, we randomly sample $l$ categories $\{c_{i_1}, c_{i_2}, \dots, c_{i_l}\}$ and input the query prompt to LLMs for generating pseudo text descriptions automatically. To filter out the unreliable sentences generated by LLMs, we re-input the generated text data into LLMs with another query template: 

\noindent\textit{PROMPT: Will the scene described in this text appear in reality? Scene:  + ``\{text\}''}.

\begin{figure}[t]
	\centering
	\includegraphics[width=\linewidth]{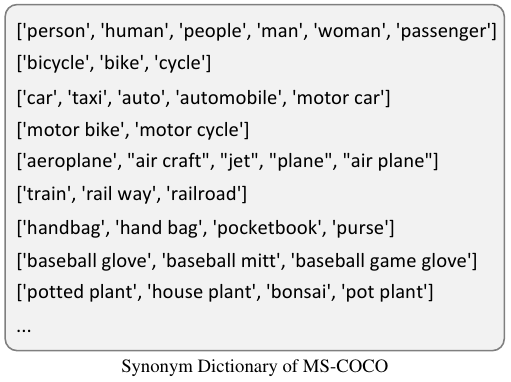}
	\caption{Illustration of synonym dictionary.}
\label{fig:synonym-word}
\end{figure}

\noindent Then we judge the reasonableness of the sentence through the output \textit{likely/unlikely}. For word-level filtered labels in input sentences, we follow the setting of TAI-DPT~\cite{TAIDPT:conf/cvpr/GuoDJBGZ23}, using NLTK\footnote{https://www.nltk.org/} to perform noun filtering. Due to each target category has synonyms with similar meanings, these synonyms also need to be mapped into the class label. Hence, we construct a synonym dictionary, which includes common synonyms of each class in the target dataset. As shown in Figure \ref{fig:synonym-word}, all the words in each row belong to the same class label. We utilize the synonym dictionary and conduct noun filtration by tokenizing and lemmatizing the words to search for sentences that contain at least one synonym name. The text data that do not match any synonym is discarded. This simple noun filtering strategy ensures that each input text contain at least one class label for prompt tuning.

\begin{figure}[t]
	\centering
	\includegraphics[width=0.85\linewidth]{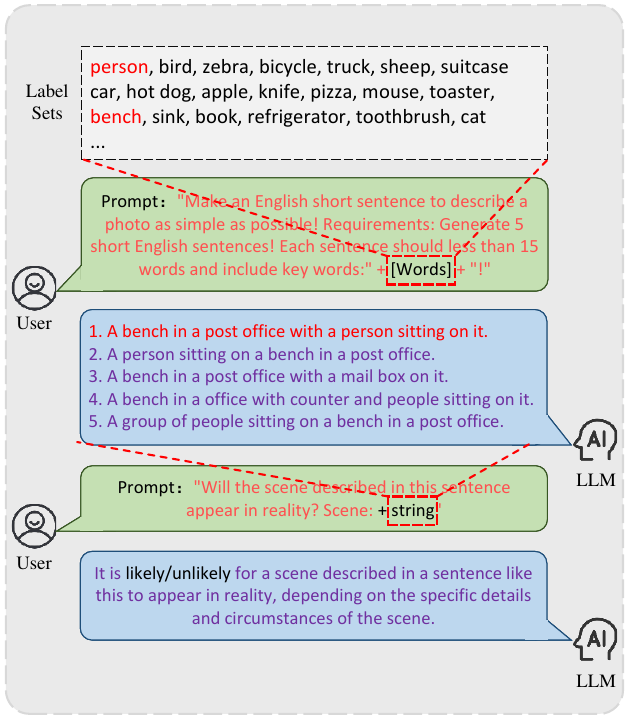}
	\caption{Pseudo text generated by LLMs. Semantically reasonable sentences are obtained via labels sampling and rationality judgment.}
\label{fig:noisy-text}
\end{figure}

In Figure~\ref{fig:noisy-text}, we provide an example for labeled text data construction. Given a set of target categories, we randomly sample several categories, such as person, bench. etc. These categories and the first prompt are constructed into a input template that can be processed by LLMs~\cite{ChatGLM:conf/iclr/ZengLDWL0YXZXTM23}, thereby generating a group of short text sentences, such as ``A bench in a post office with a person sitting on it''. We then design a rationality judgment template, and concatenate it with the text sentences obtained from the previous dialogue and input it into the LLMs again. Based on the generated results, we retain sentences with reasonable content, thereby forming noisy text training data.
 
\begin{figure*}[!ht]
\centering
\begin{tabular}{c@{ }@{ }c@{ }@{ }c@{ }@{ }c}
\begin{minipage}{0.5\linewidth}\centering
\small{(a). Bicycle} \\
\includegraphics[scale=0.52]{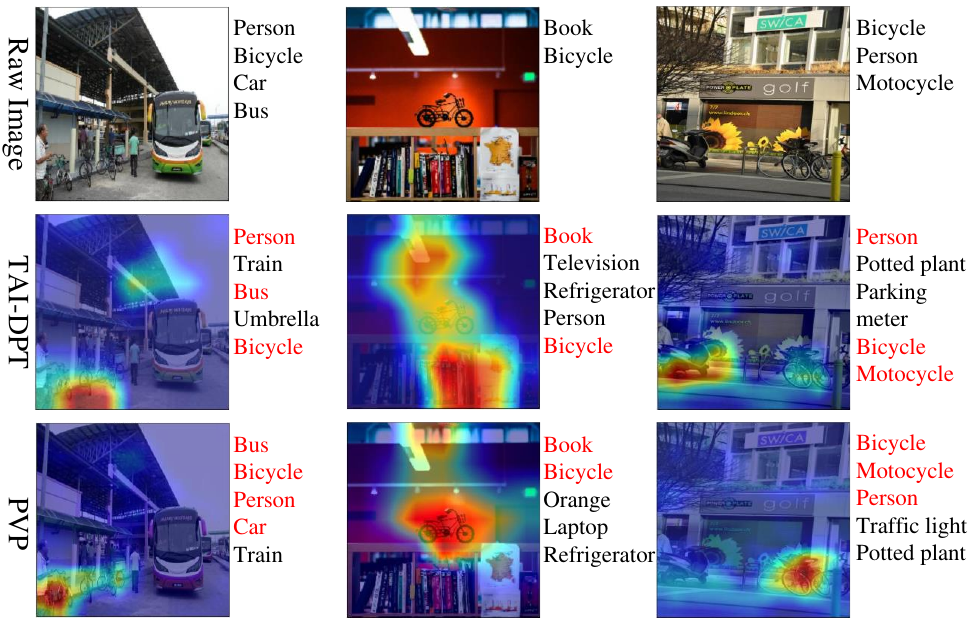}
\vspace{0.2cm}
\end{minipage} &
\begin{minipage}{0.5\linewidth}\centering
\small{(b). Hot dog} \\
\includegraphics[scale=0.52]{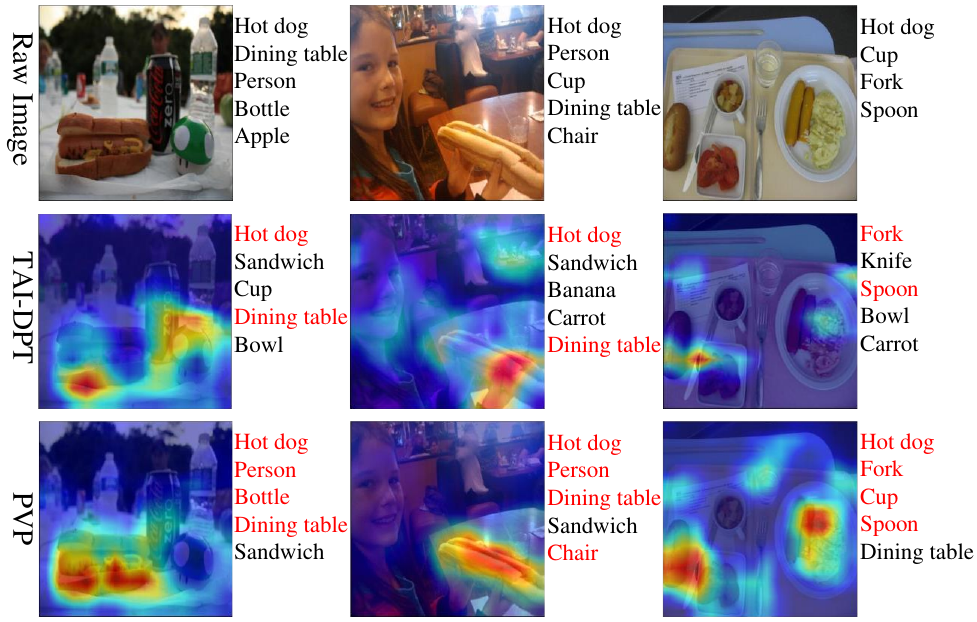}
\vspace{0.2cm}
\end{minipage} \\
\begin{minipage}{0.5\linewidth}\centering
\small{(c). Airplane} \\
\includegraphics[scale=0.52]{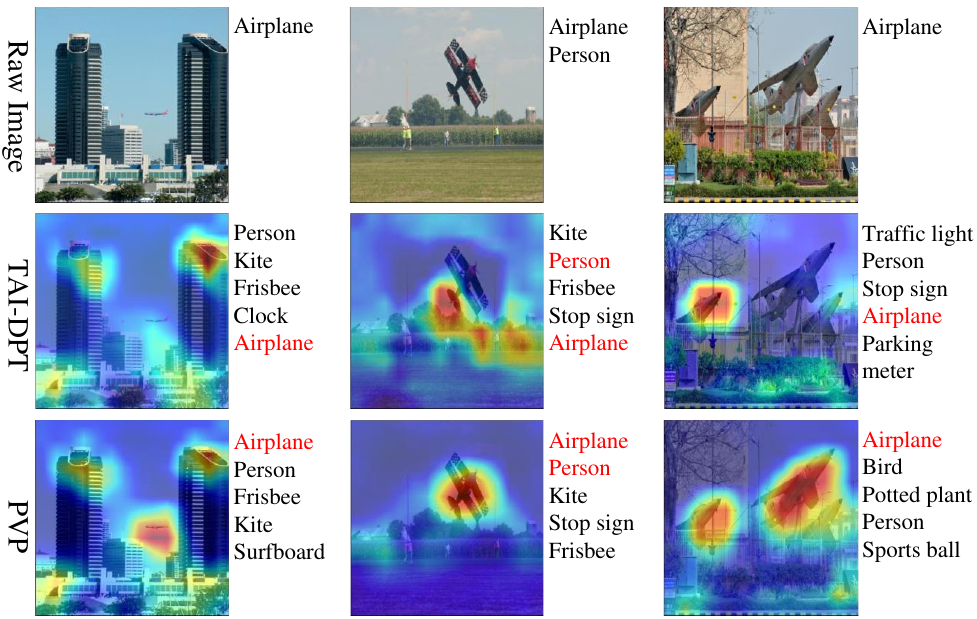}
\end{minipage} &
\begin{minipage}{0.5\linewidth}\centering
\small{(d). Television} \\
\includegraphics[scale=0.52]{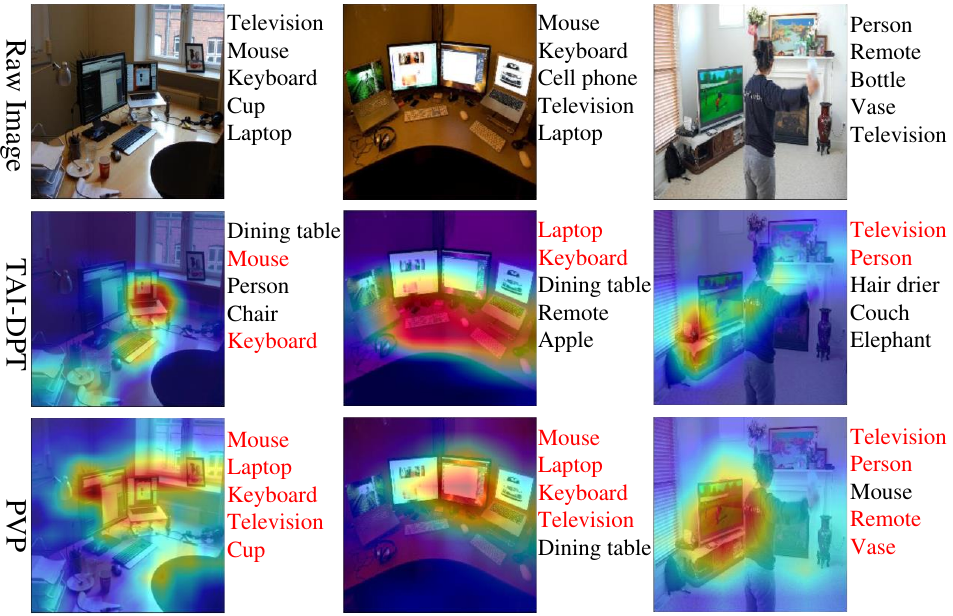}
\end{minipage}
\end{tabular}
\caption{Visualization results for categories ``bicycle'', ``hot dog'', ``airplane'' and ``television''.}
\label{fig:visualization-more}
\end{figure*}

\section{Comparison with TAI-Adapter}
Recent work TAI-Adapter~\cite{TAI-Adapter:journals/corr/abs-2312-04160} proposed to using a random perturbation mechanism to enhance the transferable capability of the adapter module. TAI-Adapter also required to use massive label visual data during training like TAI-DPT. We compare TAI-Adapter with our method in this section. 

TAI-Adapter utilizes the Vicuna-33b-1.3v~\cite{vicuna2023} to construct training text data. For fair comparison, we adopt the same LLMs to construct training text data for PVP. The mAP results\footnote{The results of TAI-Adapter are directly referred from the origin paper.} are reported in Table~\ref{tab:tai-adapter}. From Table~\ref{tab:tai-adapter}, we can find that our proposed PVP can outperform TAI-Adapter in most cases when we utilize Vicuna-33b-1.3v to generate training text data. Moreover, by comparing the mAP results based on ChatGLM and Vicuna-33b-1.3v for our proposed method, we can find that the mAP with Vicuna-33b-1.3v is higher than that with ChatGLM. The reason is that the Vicuna-33b-1.3v can extract more diverse features than ChatGLM.

\section{Visualization Results}
In this section, we illustrate more visualization results. In Figure~\ref{fig:visualization-more}, we provide some examples that are randomly selected for visualization. Specifically, we randomly select several common target categories and their corresponding raw images, visualizing the correlation between local image features and class prompt embeddings of TAI-DPT and our method PVP. For raw image, we present the ground-truth category labels. We also compute the similarities between global image feature and class prompt embeddings of TAI-DPT and PVP. And we present the top 5 categories with the highest prediction confidence from different methods. Figure~\ref{fig:visualization-more} presents the visualization results for the categories including ``bicycle'', ``hot dog'', ``airplane'' and ``television''. From Figure~\ref{fig:visualization-more}, we can see that our method can learn more diverse and comprehensive features. For the images whose shapes and attributes are different but with the same category labels, the accuracy of our method is higher than that of TAI-DPT.

\section{Further Analysis}
\subsection{Ablation Study}
\noindent{\textbf{Training Loss}}. We discuss the impact of different loss function on multi-label image recognition performance. For the contrastive training of visual and text prompts, we compare contrastive loss and ranking loss. The same loss settings are also used for the contrastive training of prompts and global text features. The results are shown in Table~\ref{tab:loss}. In Table~\ref{tab:loss}, ``CE'' and ``RL'' denote the cross-entropy and ranking loss, respectively. As shown in Table \ref{tab:loss}, applying cross-entropy loss between prompts and ranking loss between prompts and text features achieves the highest multi-label image recognition performance on VOC2007, MS-COCO, and NUSWIDE datasets.

\begin{table}[t]
\centering
\renewcommand{\arraystretch}{1.3} 
\caption{Ablation study for different loss function on all datasets.}
\label{tab:loss}\small\vspace{-5pt}
\begin{tabular}{c|c|c|c|c}
\toprule[0.8pt]
$\mathcal{L}_{visual}$/$\mathcal{L}_{text}$ & $\mathcal{L}_{vtc}$ & {VOC2007} & {MS-COCO} & {NUSWIDE} \\ \hline
CE & CE & 87.9 & 68.6 & 48.3 \\
RL & RL & 87.4 & 67.8 & 47.5 \\
CE & RL & 84.7 & 65.5 & 45.6 \\
RL & CE & \textbf{90.1} & \textbf{70.8} & \textbf{49.3} \\
\bottomrule[0.8pt]
\end{tabular}
\end{table}

\subsection{Sensitivity to Hyper-Parameters}
In this section, we present the influence of hyper-parameters, including pseudo-visual prompt size~($H$ and $W$), training epochs of the first stage. 

\begin{table}[t]
\centering
\renewcommand{\arraystretch}{1.3} 
\caption{Ablation study for the initialized size of pseudo-visual prompt on all datasets.}
\label{tab:prompt-size}
\begin{tabular}{c|c|c|c}
\toprule[0.8pt]
{PVP} & {VOC2007} & {MS-COCO} & {NUSWIDE} \\ \hline
$96\times96\times3$ & 89.4 & 69.9 & 48.5 \\
$128\times128\times3$ & 89.6 & 70.5 & 48.9 \\
$160\times160\times3$ & 89.8 & 70.7 & 48.9 \\
$224\times224\times3$ & \textbf{90.1} & \textbf{70.8} & \textbf{49.3} \\
$288\times288\times3$ & 89.9 & 70.5 & 49.2 \\
$324\times324\times3$ & 89.8 & 70.6 & 48.9 \\
\bottomrule[0.8pt]
\end{tabular}
\end{table}

\noindent{\textbf{Prompt Size.}} Pseudo-visual prompts are processed through an image encoder, with a size of $H\times W\times 3$. Therefore, we explore the impact of different prompt sizes on multi-label image classification performance. In Table~\ref{tab:prompt-size}, for VOC2007, MS-COCO, and NUSWIDE, as the prompt size gradually increases, the image recognition performance on all datasets shows a trend of first rising and then falling, reaching the best performance at the size of $224\times224\times3$. This indicates that larger prompt size can learn more extensive and diverse visual knowledge to co-learn visual and text prompts. However, an overly large prompt size will overfit the text training data, leading to weaker generalization ability in image testing, thereby affecting the performance of image recognition.

\begin{table}[!ht]
\centering
\caption{Results of the parameter analysis.}\label{tab:param-analysis}
\scalebox{1}{
\centering
\begin{tabular}{c|c|c|c|c|c}
\hline
$M$ & 8 & 12 & 16 & 20 & 24 \\\hline
mAP & 70.40 & 70.56 & 70.78 & 70.71 & 70.74 \\
\hline
\hline
$\lambda$ & 0.2 & 0.4 & 0.5 & 0.6 & 0.8 \\\hline
mAP & 70.23 & 70.62 & 70.78 & 70.75 & 70.64 \\
\hline
\end{tabular}
}
\end{table}

\noindent{\textbf{Length of text prompt.}} In our method, we set the length of the text prompt to 16 following the previous SOTA TAI-DPT. Here, we conduct several experiments to compare the influence of different prompt lengths on MSCOCO dataset. From Table 8, the prompt length (parameter \textit{M}) ranges from 8 to 24, we can see the performance of different prompt lengths is similar to others, and the larger size of the prompt can not further improve the image classification results.

\noindent{\textbf{Weight of dual-adapter and CLIP.}} Dual-adapter is designed to both learn the new knowledge of downstream datasets and maintain the origin knowledge of the pretrained CLIP. Therefore, we explore the weight of dual-adapter and origin CLIP to evaluate the importance of both the downstream MSCOCO dataset and CLIP. From Table 8, the weight, denoted as $\lambda$, ranges from 0.2 to 0.8, PVP achieves the best performance with the $l\lambda$ being 0.5, demonstrating the knowledge information of MSCOCO and origin CLIP are equally important.

\begin{table}[!ht]
    \centering
    \caption{Analysis for $\gamma$, $\eta$, and $\nu$.}
    \resizebox{0.5\columnwidth}{!}{
    \begin{tabular}{c|c|c|c}
    \hline
    $\gamma$ & $\eta$ & $\nu$ & {MSCOCO} \\
    \hline
    1 & 1 & 1 & 70.78 \\
    1 & 2 & 3 & 70.62 \\
    1 & 3 & 2 & 70.63 \\
    2 & 1 & 3 & 70.76 \\
    2 & 3 & 1 & 70.80 \\
    3 & 1 & 2 & 70.65 \\ 
    3 & 2 & 1 & 70.69 \\ 
    \hline
    \end{tabular}
    \label{tab:ablation-study}
    }
\end{table}

\noindent{\bf Weights of Loss:} Here, we provide the experimental results with different loss weights. We first rewrite the objective function as $\mathcal{L}=\gamma\mathcal{L}_{vtc}+\eta\mathcal{L}_{visual}+\nu\mathcal{L}_{text}$. The experimental results are shown in Table~\ref{tab:ablation-study}. From Table~\ref{tab:ablation-study}, we can see that the mAP almost remains unchanged with different values of weights. We will discuss the influence of the weights within a larger range in the final version.

\begin{figure}[t]
	\centering
	\includegraphics[width=0.7\linewidth]{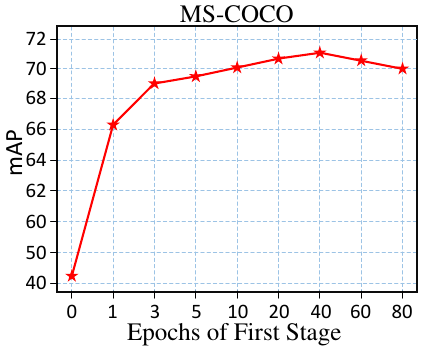}
	\caption{Ablation study for the training epochs of pseudo-visual prompt learning stage on MS-COCO dataset.}
\label{fig:epochs}
\end{figure}

\noindent{\textbf{Training Epochs of The First Stage.}} Our method consists of two stages: pseudo-visual prompt learning and transferable visual and text prompts co-learning. We explore how the epochs of pseudo-visual prompt learning in the first stage affect the multi-label image recognition performance in the second stage. Figure~\ref{fig:epochs} shows that the longer the pseudo-visual prompts are learned, the greater the improvement in image recognition performance on MS-COCO dataset. Moreover, the case where epoch equals 0 denotes that pseudo-visual prompt and text prompt co-learning is performed directly without the pseudo-visual prompt learning in the first stage. This result indicates that the mAP results will be hindered if we only perform transferable prompt co-learning without pseudo-visual prompt learning.

